# A new muscle fatigue and recovery model and its ergonomics application in human simulation

Liang MA[1], Damien CHABLAT[1], Fouad BENNIS [1], Wei ZHANG [2], François GUILLAUME [3]

(1): IRCCyN - 1, rue de la Noë - BP 92 101 - 44321 Nantes CEDEX 03, France
E-mail: {liang.ma, damien.chablat, fouad.bennis}@irccyn.ec-nantes.fr

(2): Department of Industrial Engineering, Tsinghua University, 100084, Beijing, China
E-mail: zhangwei@tsinghua.edu.cn

(3): EADS Innovation Works, 12, rue Pasteur – BP 76, 92152 Suresnes Cedex - FRANCE
E-mail: francois.guillaume@eads.net

**Abstract:** Although automatic techniques have been employed in manufacturing industries to increase productivity and efficiency, there are still lots of manual handling jobs, especially for assembly and maintenance jobs. In these jobs, musculoskeletal disorders (MSDs) are one of the major health problems due to overload and cumulative physical fatigue. With combination of conventional posture analysis techniques, digital human modelling and simulation (DHM) techniques have been developed and commercialized to evaluate the potential physical exposures. However, those ergonomics analysis tools are mainly based on posture analysis techniques, and until now there is still no fatigue index available in the commercial software to evaluate the physical fatigue easily and quickly. In this paper, a new muscle fatigue and recovery model is proposed and extended to evaluate joint fatigue level in manual handling jobs. A special application case is described and analyzed by digital human simulation technique.

## 1- Introduction

Automation in industry has been increased in recent years and more and more efforts have been made to achieve efficient and flexible manufacturing. However, manual work is still very important due to increase of customized products and human's capability of learning and adapting [FM1]. Musculoskeletal disorder (MSD) is the injuries and disorders to muscles, nerves, tendons, ligaments, joints, cartilage and spinal discs [MR1]. From the report of Health, Safety and Executive [H1] and the report of Washington State Department of Labor and Industries [S1], over 50% of workers in industry have suffered from musculoskeletal disorders, especially for manual handling jobs. According to the analysis in Occupational Biomechanics [CA1], "Overexertion of muscle force or frequent high muscle load is the main reason for muscle fatigue, and furthermore, it results in acute muscle fatigue, pain in muscles and severe functional disability in muscles and other tissues of the human body". Hence, it is very important for ergonomists to find an efficient method to assess the extent of various physical exposures on muscles and to predict muscle fatigue in the work design stage.

In order to assess physical risks to MSDs, there are several posture based ergonomics tools for posture analysis, such as Posturegram, Ovako Working Posture Analyzing System (OWAS), Posture Targeting and Quick Exposure Check for work-related musculoskeletal risks (QEC). In spite of these general posture analysis tools, some special tools are designed for specific parts of the human body. Rapid Upper Limb Assessment (RULA) is designed for assessing the severity of postural loading for the upper extremity. The similar systems include HAMA (Hand-Arm-Movement Analysis), PLIBEL (method for the identification of musculoskeletal stress factors that may have injurious effects) [SH1]. Similar to these methods for posture analysis, there is one tool available for fatigue analysis and that is muscle fatigue analysis (MFA). This technique was developed to characterize the discomfort described by workers on automobile assembly lines and fabrication tasks [R1]. In this method, each body part is scaled into four effort levels according to its working position, duration of the effort, and frequency. The combination of the three factors' levels can determine a "priority to change" score. The task with a high priority score needs to be analyzed and redesigned to reduce the MSD risks [SH1, R2].

After listing these available methods, physical exposure to MSD can be evaluated with respect to its intensity (or magnitude), repetitiveness, and duration [LB1]. However, there are still several limitations with the traditional methods. First, the evaluation techniques lack precision and their reliability of the system is a problem for assessing the physical exposures due to their intermittent recording procedures [B1]. Second, most of the traditional methods have to be carried out on site. Therefore, there is no immediate result from the observation. It is also time consuming for later analysis. Furthermore, subjective variability can influence the evaluation results when using the same observation methods for the same task [DH1].

In order to evaluate the human work condition objectively and quickly, digital human techniques have been developed to facilitate the ergonomic evaluation, such as Jack [BP1], ErgoMan [SL1], 3DSSPP [C1], Santos [V1]. These techniques have been used in the fields of automotive, military, and aerospace. These human modelling and simulation tools provide mainly visualization information about body posture; accessibility and field of view [DH1]. Combining Digital Mock-Up (DMU) with digital human models (DHM), the simulated human associated with graphics could supply visualization of the work design, and it could decrease the design time and enhance the number and quality of design options that could be rapidly evaluated by the design analysts [C2]. Traditional posture analysis tools have been integrated into these simulation tools for computerization. For example, in 3DSSPP, in CATIA, and in other simulation tools, RULA, OWAS and some other posture analysis tools have been integrated as a module to evaluate the postures in design stage. In these digital human simulation tools, it is possible to generate the motion for certain task, and the load of each key joint and even each muscle can be determined and simulated. In [JJ1], a method to link virtual environment (Jack) and a quantitative ergonomic analysis tool (RULA) for occupational ergonomics studies was developed. This framework verified the conception of evaluating ergonomics study in real time manner by obtaining human motion from motion capture system.

However, even today, there is still no effective method in these digital human modelling and simulation tools to predict human motion with consideration of muscle fatigue, and there is still no fatigue evaluation tool integrated in these human simulation tools. Therefore, it is necessary to develop the muscle fatigue model and then integrate it into the virtual human software to evaluate muscle fatigue and specifically analyze the physical work, and even predict the human motion by minimizing the fatigue.

Several muscle fatigue models and fatigue indices have been proposed in the literature. In a series of publications [WD1, DW1, DW2, and DW3], a new muscle fatigue model based on $Ca^{2+}$ cross-bridge mechanism was verified by stimulation experiments. This model based on the physiological mechanism seems too complex for ergonomic application due to its large number of variables. Another muscle fatigue model [GM1] based on force-pH relationship was obtained by curve fitting of the pH level with time in the course of stimulation and recovery. Komura et al. [KS1, KS2] have used this model in computer graphics to visualize the muscle capacity. However, in this pH muscle fatigue model, all the influences on fatigue from physical aspects are not considered. Rodriguez proposed a half-joint fatigue index in the literature [RB1, RB2, and RB3] based on mechanical properties of muscle groups. This fatigue model was used to calculate the fatigue at joint level, and the fatigue level is expressed as the actual holding time normalized by maximum holding time of the half-joint. The maximum holding time equation of this model was from static posture analysis and it is mainly suitable for evaluating static postures. Because of these limitations in current existing fatigue models, a new simple model is necessary to evaluate the fatigue.

In this paper, we are going to present a new framework to evaluate the manual handling jobs objectively and quickly in a virtual environment. In this framework, a new muscle fatigue and recovery model is integrated to evaluate the fatigue and decide the work-rest schedule. A simplified geometrical and biomechanical model of arm is constructed to calculate the load of each joint using inverse dynamics. A special case in EADS is used to evaluate the fatigue of the manual handling job.

## 2- Framework for the fatigue analysis

In order to evaluate manual handling work objectively and effectively, a framework based on virtual reality technique is graphically presented in Figure 1.

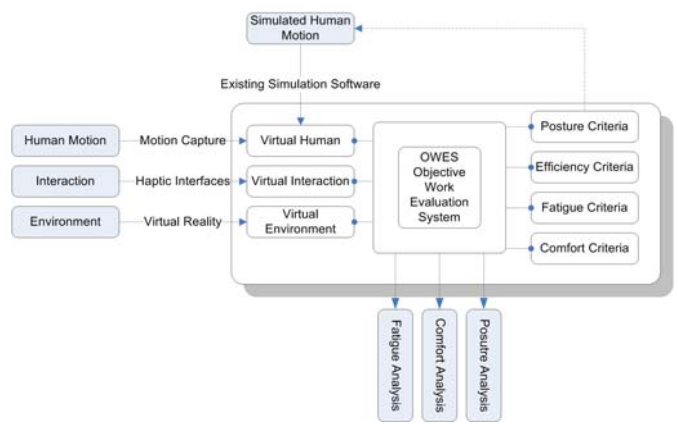

**Figure 1: Framework for objective work evaluation system**

The overall function of the framework is to field-independently evaluate the difficulty of human mechanical work including fatigue, comfort and other aspects. The framework consists of three main modules: virtual environment module, data collection module, and evaluation module.

The module of virtual environment technique and virtual human technique is used to provide the virtual working environment and to avoid field-dependent work evaluation. Based on VE, immersive work simulation system is constructed to provide the virtual working environment. Virtual human is modelled and driven by the motion data to generate the manual handling job in the virtual environment. Another component, haptic interface is used to enable the interactions between the worker and virtual environment.

Data collection module is responsible for obtaining all the necessary information for further data processing. From the introduction part, necessary information for evaluating dynamic manual handling jobs consists of motion, forces and personal factors. To achieve the motion data, motion capture technique can be applied to achieve the motion information with individuality. Nevertheless, the motion information can also be achieved from some existing human simulation tools. Personal factors can be obtained from anthropometry database or measurements. The forces can be measured by force measurement devices or known external loads.

The evaluation module takes all the input data to evaluate the manual operation. In this module, evaluation criteria of all the aspects of the manual operation are predefined in the framework, such as posture analysis criteria, fatigue criteria and discomfort criteria. With these criteria, different aspect can be evaluated by processing the input data.

The detailed technical analysis of the framework was presented in the literature [MB1], and here we just make a brief introduction to its work flow. In this framework, at first the manual handling operation is carried out in the virtual environment module. Virtual working environment is provided for visualization. Human's motion in a manual handling operation is either captured from motion capture system or simulated using human simulation software. The motion information combined with the interaction information with the virtual environment is collected and further processed in the objective work evaluation module. In this module, with the predefined criteria, different aspects of the manual operation can be evaluated. The evaluation results can be used for further improvement of the work design.

## 3- Muscle fatigue and recovery model

Muscle fatigue is defined as the point at which the muscle is no longer able to sustain the required force or work output level [V1]. In order to evaluate the muscle fatigue during a manual handling operation, a new muscle fatigue and recovery model was developed based on muscle motor mechanisms pattern, and the details are presented in this section. At first, the parameters in this muscle fatigue and recovery model are listed in Table 1.

| Parameters | Unit | Description |
|---|---|---|
| $U$ | - | Fatigue index |
| $MVC$ | N | Maximum voluntary contraction |
| $F_{cem}$ | N | Muscle force capacity at time instant $t$ |
| $F_{load}$ | N | Muscle load at time instant $t$ |
| $\Gamma_{max}$ | Nm | Maximum joint strength |
| $\Gamma_{cem}$ | Nm | Joint strength at time instant $t$ |
| $\Gamma$ | Nm | Torque at the joint at time instant $t$ |
| $k$ | min$^{-1}$ | Fatigue ratio, equals to 1 |
| $R$ | min$^{-1}$ | Recovery ratio, equals to 2.4 |
| $t$ | min | Time |

**Table 1: Parameters in muscle fatigue and recovery model**

### 3.1- Muscle fatigue model

The muscle fatigue model is based on motor mechanism pattern of muscles. A muscle consists of many motor units. Each motor unit has different force generation capability, and different fatigue and recovery properties. In general, there are three types in the muscle: type I is slow-twitch motor units with small force generation capability and low conduction velocity, but a very high fatigue resistance; type II b is of fast-twitch speed, high force capacity, but fast fatigability; type II a, between type I and type II b, has a moderate force capacity and moderate fatigue resistance. The sequence of recruitment is in the order of: I → II a → II b [V1]. For a specified muscle, larger $F_{load}$ means more type II motor units are involved to generate the force. As a result, the muscle becomes fatigued more rapidly, as expressed in Eq. (2). $F_{cem}$ represents the non-fatigue motor units of the muscle. In the process of force generation, the amount of non-fatigued type II motor units gets smaller and smaller due to fatigue, while the number of the type I motor units remains almost the same due to their high fatigue resistance, and the decrease of $F_{cem}$ with time becomes slower, as expressed in Eq. (2) by term $F_{cem}(t)/MVC$. This muscle fatigue model has been mathematically validated by comparing 24 existing static endurance time models listed in [EK1] and 3 dynamic models in [LB2, FT1, DW3] in [MC1]. The validation result proves that this model is capable for muscle fatigue evaluation.

$$\frac{dU}{dt} = \frac{MVC}{F_{cem}} \frac{F_{load}}{F_{cem}} \quad (1)$$

$$\frac{dF_{cem}}{dt} = -k \frac{F_{cem}}{MVC} F_{load} \quad (2)$$

### 3.2- Muscle recovery model

This model (Eq.(3)) is developed based on recovery models mentioned in the literature [WF1, CN1]. This model can also be explained by muscle motor mechanism pattern. $(MVC-F_{cem})$ represents the fatigued motor units in the muscle. The recovery rate from fatigue muscle motor units is assumed to be constant 2.4 [LB2, WF1], in symbol $R$.

$$\frac{dF_{cem}}{dt} = R(MVC - F_{cem}) \quad (3)$$

Therefore, the $F_{cem}$ can be determined by Eq. (4)

$$F_{cem}^t = MVC + (F_{cem}^0 - MVC)e^{-Rt} \quad (4)$$

With this recovery model, the recovery time from a certain fatigue level $F^0_{cem}$ to $p$ percentage of $MVC$ can be determined by Eq. (5).

$$t = -\frac{1}{R}\left(\frac{(p-1)MVC}{F_{cem}^0 - MVC}\right) \quad (5)$$

### 3.3- Extension of this model to joint level

The muscles attached around a joint are responsible for generating torque to move the joint or keep it stable for maintaining the external load. There are several muscle engaged in generating a simple movement of the arm. Mathematically, to determine the efforts of each muscle involved in the movement is an underdetermined problem, so it is difficult to determine the actual load of each muscle. Although some optimisation methods have been created to solve force distribution problem in muscle levels, it is not easy to achieve the accurate result for each individual muscle. However, according to inverse dynamics, it is accurate enough to calculate the torque of each joint. And meanwhile, in ergonomics application, the analysts do often evaluate the physical exposures in joint level. $MVC$ is sometimes defined in the literature [ME1] as joint strength. For this reason, this muscle fatigue and recovery model is extended to evaluate the fatigue and joint level by simply replacing the parameters in the muscle model. *MVC* is replaced by the maximum joint strength $\Gamma_{max}$. $F_{cem}$ is replaced by current joint strength with time $\Gamma_{cem}$, and $F_{load}$ is replaced by the joint load torque $\Gamma$. The other parameters are kept the same in the model. The extension of the model is also mathematically validated by comparing the existing models in [MC1].

The muscle model fatigue and recovery model can be used to analyze the performance of an individual muscle. The extended model is available to analyze muscle groups performance, in other words, reduction of joint strength in a continuous working process.

## 4- Application of the fatigue model

### 4.1- Special application cases in EADS

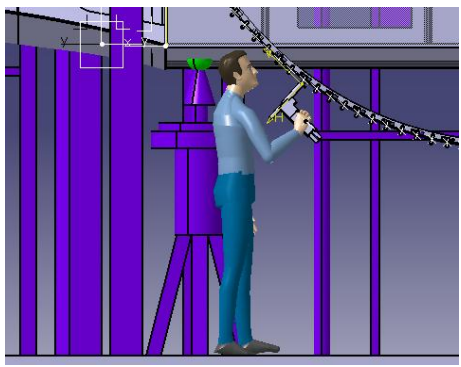

**Figure 2: Drilling task in EADS field application**

In our research project, the application case is junction of two fuselage section with rivets from the assembly line of a virtual aircraft. One part of the job consists of drilling holes all around the section. The properties of this task can be described in natural language as: drilling holes around the fuselage circumference. The number of the holes could be up to 2000 under real work conditions. The drilling machine has a weight around 5 kg, and even up to 7 kg in the worst condition with consideration of the pipe weight. The drilling force applied to the drilling machine is around 49N. In general, it takes 30 seconds to finish a hole. The drilling operation is graphically shown in Figure 2.

In this application case, there are several ergonomics issues and several physical exposures contribute to the difficulty and penalty of the job. It includes posture, heavy load from the drilling effort, the weight of the drilling machine, and vibration. Muscle fatigue is mainly caused by the load on certain postures, and the vibration might result in damage to some other tissues of arm. To maintain the drilling work for a certain time, the load could cause fatigue in elbow, shoulder, and lower back. In this paper, the analysis is only carried out to evaluate the fatigue of right arm in order to verify the conception of the framework. The vibration is excluded from the analysis. Further more, the external loads are divided by two in order to simplify the calculation, for two arms are usually engaged in drilling operation.

### 4.2- Geometrical modelling of arm

According to the new fatigue model, it is important to calculate the joint torques of human; therefore, geometrical model of the right arm is developed using the modified Denavit-Hartenberg (DH) notation methods [KK1] to describe the geometric structure of the right arm. In modified DH notation system, four parameters are used to describe the transformation between two Cartesian coordinates in Figure 3.

1. $\alpha_j$: angle between axes $Z_{j-1}$ and $Z_j$ around the axis $X_{j-1}$.
2. $d_j$: distance between axes $Z_{j-1}$ and $Z_j$ along the axis $X_{j-1}$.
3. $\theta_j$: angle between axes $X_{j-1}$ and $X_j$ around the axis $Z_j$
4. $r_j$: distance between axes $X_{j-1}$ and $X_j$ along the axis $Z_j$.

From anatomic, the shoulder joint allows the movement as a sphere joint in flexion and extension, adduction and abduction, and supination and pronation directions. Elbow joint is able to move in flexion and extension direction and supination and in pronation direction.

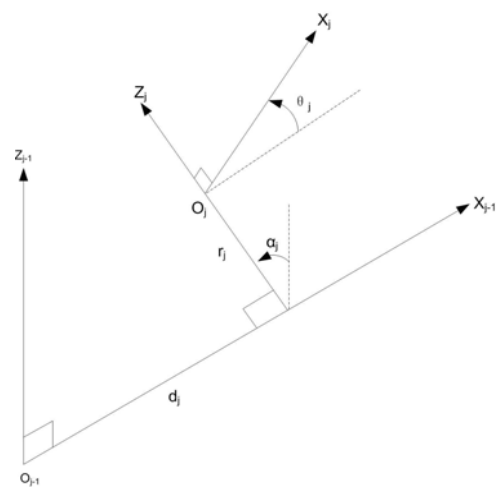

**Figure 3: Modified Denavit-Hartenberg notation system**

The shoulder complex is separated into 3 rotational joints and the elbow joint is separated into 2 rotational joints shown in Figure 4. Each joint has its own joint coordinate system defined in DH notation system, and the joint can only rotate around its Z-axis within rotation limits. The anatomical function of each joint is explained in Table 2. The parameters in modified DH notation system are listed in Table 3, and the transformation matrix between current joint coordinate to precedent joint coordinate is Eq. (6). The right arm is geometrically represented by a chain of rotational joints, by a general vector $q=[q_1,q_2,q_3,q_4,q_5]$. Each element $q_i$ represents the rotation angle around the Z-axis in $R_i$. Once the geometrical configuration $q$ is given, the posture of the arm can be fixed.

| Joints | Description |
|--------|-------------|
| 1 | Flexion and extension of shoulder joint |
| 2 | Adduction and abduction of shoulder joint |
| 3 | Supination and pronation of upper arm |
| 4 | Flexion and extension of shoulder joint |
| 5 | Supination and pronation of upper arm |

**Table 2: Geometrical parameters for modelling right arm**

| Joint | σ | α | d | r | θ | $\theta_{ini}$ |
|---|---|---|---|---|---|---|
| 1 | 0 | -π/2 | 0 | 0 | $\theta_1$ | -π/2 |
| 2 | 0 | -π/2 | 0 | 0 | $\theta_2$ | -π/2 |
| 3 | 0 | -π/2 | 0 | -RL3 | $\theta_3$ | -π/2 |
| 4 | 0 | -π/2 | 0 | 0 | $\theta_4$ | 0 |
| 5 | 0 | π/2 | 0 | 0 | $\theta_5$ | 0 |

**Table 3: Geometrical parameters for modelling right arm**

$$^{j-1}T_j = \begin{bmatrix} \cos\theta_j & -\sin\theta_j & 0 & d_j \\ \cos\alpha_j \sin\theta_j & \cos\alpha_j \cos\theta_j & -\sin\alpha_j & -r_j \cos\alpha_j \\ \sin\alpha_j \sin\theta_j & \sin\alpha_j \cos\theta_j & \cos\alpha_j & r_j \sin\alpha_j \\ 0 & 0 & 0 & 1 \end{bmatrix} \quad (6)$$

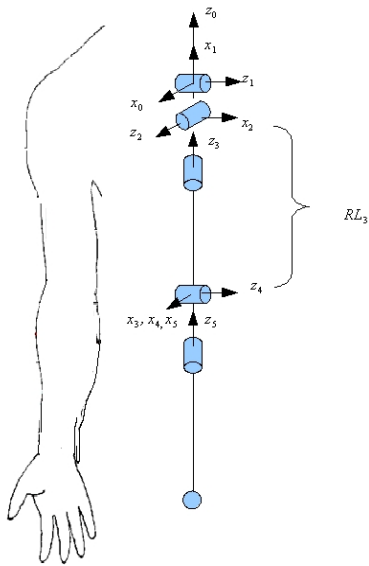

**Figure 4: Geometrical model of the arm**

4.3- Dynamical parameters of arm

| Parameters | Unit | Description |
|---|---|---|
| $M$ | kg | mass of the virtual human |
| $H$ | m | height of the virtual human |
| $m$ | kg | mass of the segment |
| $f$ | - | subscript for forearm |
| $u$ | - | subscript for upper arm |
| $I_G$ | - | moment of inertia of the segment |
| $h$ | m | length of the segment |
| $r$ | m | radius of the segment |

**Table 4: Dynamic parameters for modelling the right arm**

The arm is segmented into two parts: upper arm and forearm (hand included). Each part of the arm is simplified to a cylinder form and assumed a uniform distribution of density in order to calculate its moment of inertia. The weight and dimensional information of the arm can be achieved from anthropometry in occupational biomechanics [CA1] by Eq. (7) and Eq. (8), with $M$ as weight of the digital human and $H$ as height of digital human. Once the weight $m$ and cylinder radius $r$ and height $h$ are known, its inertia moment around its long axis can be determined by a diagonal matrix in Eq. (9).

$$\begin{cases} m_f = 0.451 \times 0.051 M \\ m_u = 0.549 \times 0.051 M \end{cases} \quad (7)$$

$$\begin{cases} h_f = 0.146 H \\ r_f = 0.125 h_f \\ h_u = 0.186 H \\ r_u = 0.125 h_u \end{cases} \quad (8)$$

$$I_G = \text{diag}\left( \frac{mr^2}{4} + \frac{mh^2}{12}, \frac{mr^2}{4} + \frac{mh^2}{12}, \frac{mr^2}{2} \right) \quad (9)$$

In our case, a digital human weighted as 70 kg and with a height as 1.70m is chosen to calculate those parameters of the right arm.

4.4- Calculation of internal joint forces and torques

The source of the external load is original from two parts: the gravity of the drilling machine with direction vertical down, and the drilling effort in direction of the hole. The forces and torques at each joint can be calculated following Newton-Euler inverse dynamic methods mentioned in book [KD1]. At the end, the forces and torques are projected into general joint coordinates to calculate the effort generating the corresponding movement of the joint.

4.5- Fatigue evaluation of the joints

As mentioned in the fatigue model, it is necessary to find out the joint strength in order to evaluate the joint fatigue. The standard strength data of shoulder and elbow can be obtained from the occupational biomechanics [CA1]. The flexion strength of shoulder and elbow are mainly depending on gender and flexion angles of the arm. In this case, the $q_1$ and $q_4$ are used as variables to calculate the joint strength. The result of the joint strength is the mean value $\Gamma_j$ of the population and its standard deviation $\sigma_j$. In order to analyze the compatibility of the population, 95% ($\Gamma_j \pm 2\sigma_j$) population is taken into consideration in our analysis. As an example, the elbow flexion strength for the 95% male adult population is graphically shown in Figure 5. Two geometrical variables, flexion angle of elbow and flexion angle of shoulder, are used to calculate the elbow flexion joint strength. It is obvious that different geometrical configuration determines different flexion joint strength and that the variation of the strength among the population is quite large.

The joint strength in a given geometrical configuration can be calculated, and then with the new fatigue model, the reduction of the joint strength can be evaluated.

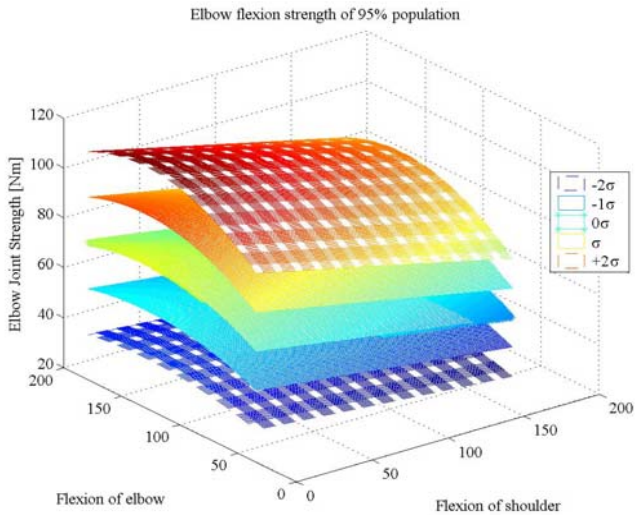

**Figure 5: Flexion elbow joint strength within the joint limits.**

### 5- Results and Discussion

#### 5.1- Results

##### 5.1.1- Endurance Time for continuous work

| Height | 1,70 m | Weight | 70 kg |
|---|---|---|---|
| $\Delta q_1$ | -30° | $\Delta q_4$ | -90° |
|  | Mean | Std. Deviation | 95% Population |
| $\Gamma_{shoulder}$ [Nm] | 75.620 | 17.476 |  |
| $\Gamma_{elbow}$ [Nm] | 75.141 | 18.470 |  |
| Extenral Load | $\Gamma_1$ (Nm) | $\Gamma_4$ (Nm) |  |
| 2.5kg | 23.043 | 7.394 |  |
| 3.5kg | 26.873 | 9.672 |  |

**Table 5: Initial parameters and joints flexion strength under geometrical configuration $α_E$ =90° and $α_S$ =90°**

With the new fatigue model, a continuous work procedure is evaluated under a geometrical configuration of the arm listed in Table 5. $\Delta q_1$ means the flexion of the shoulder, and $\Delta q_4$ means the flexion of the shoulder. The sign of both variables indicates the rotation direction around its Z-axis. With this geometrical configuration, the strength and variation of the joint can be determined, and they are also listed in Table 5. Different external load generates different torque in flexion joints.

The endurance time for the static posture is listed in Table 6. Under the same geometrical configuration, different load influences the endurance time. For shoulder, even the difference of the shoulder load is about 4 *Nm*, but it could decrease almost one forth of endurance time. The higher external load is, the shorter endurance time for maintaining the job is. It is quite clear that different capacity of the population can do the same task with quite different performance. It varies from 60 s to 450 s for drilling a same hole until exhausted stage. For maintaining the posture, the shoulder and elbow have different endurance time. For the overall work capacity evaluation, the minimum capacity is used to avoid any injury on human body. From the last two rows of Table 6, the number of holes which the worker is able to drill in a continuous working procedure is shown. Using our fatigue index, the fatigue of each joint is also evaluated. For drilling only one hole in 30 seconds, the maximum fatigue index occurs at the negative side of the population in the shoulder joint (0.330).

| External Load | -2σ | -σ | 0 | -2σ | +2σ |
|---|---|---|---|---|---|
| 2.5 kg, Shoulder [s] | 60.155 | 140.125 | 233.984 | 338.456 | 451.520 |
| Us of 30 s * | 0.283 | 0.198 | 0.152 | 0.124 | 0.104 |
| 2.5 kg, Elbow [s] | 509.083 | 936.582 | 1413.831 | 1928.300 | 2472.535 |
| Ue of 30 s * | 0.097 | 0.065 | 0.049 | 0.039 | 0.033 |
| 2.5 kg Holes | 2 | 5 | 8 | 11 | 15 |
| 3.5 kg, Shoulder [s] | 37.623 | 100.198 | 174.683 | 258.268 | 349.221 |
| Us of 30 s | 0.330 | 0.231 | 0.178 | 0.144 | 0.122 |
| 3.5 kg, Elbow [s] | 325.501 | 621.517 | 955.564 | 1318.062 | 1703.315 |
| Ue of 30 s | 0.127 | 0.085 | 0.064 | 0.052 | 0.043 |
| 3.5 kg Holes | 1 | 3 | 6 | 9 | 11 |
| External load | Recovery time for 30 s drilling work [s] | | | | |
| 3.5 kg Shoulder | 83.542 | 75.758 | 69.815 | 65.011 | 60.981 |
| 3.5 kg Shoulder | 61.945 | 52.576 | 45.774 | 40.432 | 36.033 |
| 2.5 kg Elbow | 80.243 | 72.301 | 66.270 | 61.412 | 57.343 |
| 2.5 kg Elbow | 55.584 | 46.101 | 39.240 | 33.863 | 29.439 |
| *Us, Ue : Fatigue index of shoulder and elbow | | | | | |

**Table 6: Endurance time [s] and fatigue index U of shoulder and elbow flexion joints under continuous working condition**

From Figure 6 to Figure 9, the reduction of the joint strength during the operation is graphically presented. In a continuous static posture holding procedure, there is no recovery of the joint strength. The joint strength decreases with time.

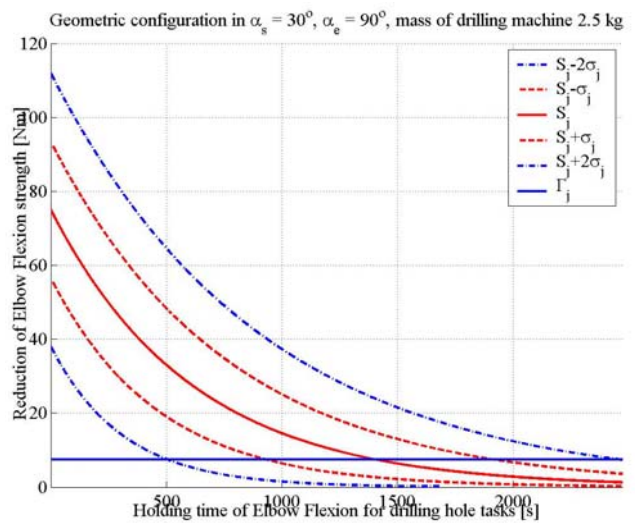

**Figure 6: Reduction of the elbow strength while holding a drilling machine weighted as 2.5 kg**

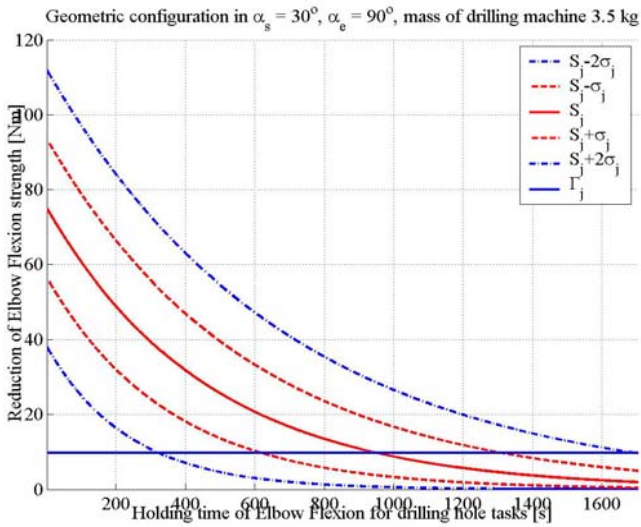

**Figure 7: Reduction of the elbow strength while holding a drilling machine weighted as 3.5 kg**

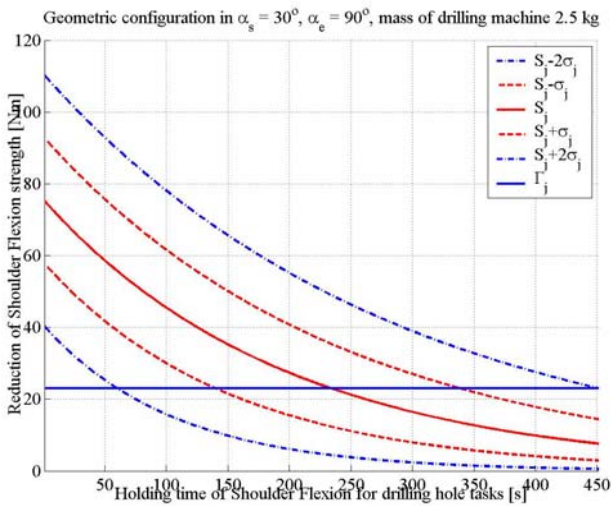

**Figure 8: Reduction of the shoulder strength while holding a drilling machine weighted as 2.5 kg**

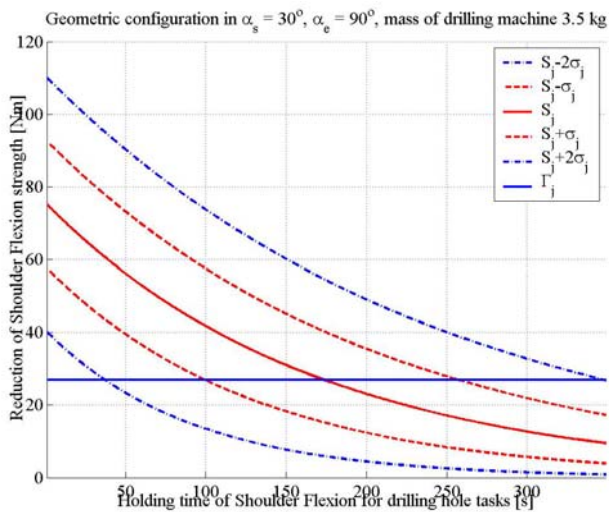

**Figure 9: Reduction of the shoulder strength while holding a drilling machine weighted as 3.5 kg**

### 5.1.2- Influence of Recovery

Work-rest schedule is very important in ergonomics application. Combining fatigue and recovery model can determine the work-rest schedule. Different work cycle results in different fatigue evaluation results. In our case, two working cycle are evaluated. One is drilling a hole in 30 s and recovery 30s in Figure 10, and another one is 30s drilling and 60s recovery in Figure 11. From previous analysis, we take the 3.5kg and shoulder joint for demonstrate the influence of recovery period. It is obvious that the longer the rest period is, the better the joint strength can be recovered. Sufficient recovery time can maintain the worker's physical capacity for quite a long time; but insufficient recovery time might cause cumulative fatigue in the joint. In Figure 10, cumulative fatigue during the working procedure can be indicated by the reduction of the joint strength. And in rest time 60 s, the joint strength can be recovered during the rest period to maintain the job. Once the requirement of the joint strength is over the capacity; the overexertion might cause MSD in human body. It should be mentioned that in actual work; there are lots of influencing factors affecting the recovery procedure, and the recovery ratio is changed individually. According to [LB2, WF1], $R$ is set as 2.4 min$^{-1}$ for 50% population to determine the work-rest schedule.

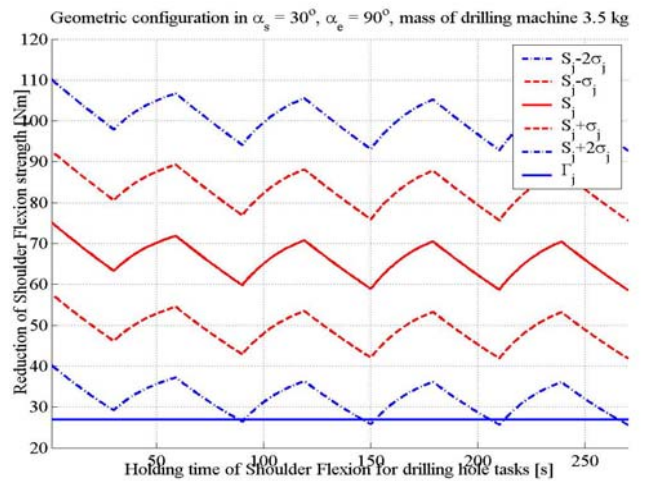

**Figure 10 : Recovery 30 s after drilling a hole**

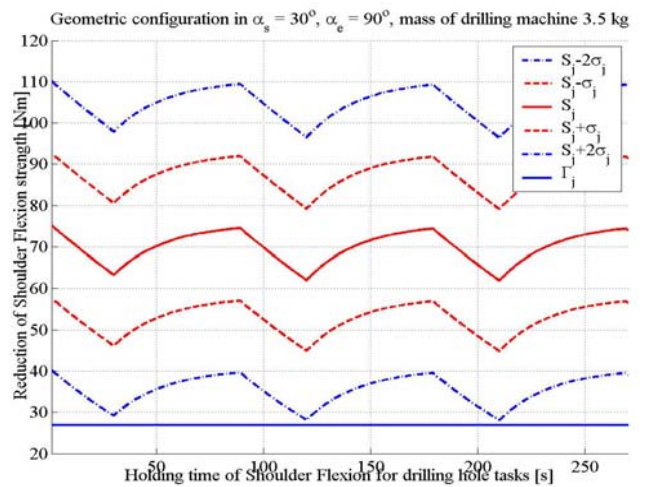

**Figure 11: Recovery 60 s after drilling a hole**

### 5.1.3- With consideration of discomfort

In fact, fatigue is not a single aspect in ergonomics analysis. There are some other factors influencing the actual operation of the worker, such as joint discomfort, so the posture prediction is a multi-objective optimisation problem.

In our application, fatigue and discomfort are combined into Eq. (10) to convert the multi-objective function into a single objective function for posture prediction.

The fatigue index (stress index) is expressed by the summation of the relative joint load. In paper [YM1], a discomfort index is proposed as an objective to predict human motion and it is taken into our framework to evaluate the joint discomfort. This discomfort index from Eq. (12) to Eq. (15) estimates the comfort of the joint by comparing its current position with its upper limitation, lower limitation and its neutral position. The most comfortable position is in neutral position of the joint.

In our case, right shoulder and the hole are predefined in the same horizontal line in sagittal plane. Therefore, different postures need to be adjusted to adapt to the variation of the distance. Different posture causes different fatigue and different discomfort. Therefore, using the discomfort index and stress index in Eq. (11) and Eq (12), an optimal posture can be found to balance the requirement of fatigue and discomfort. The results are shown in Figure 12.

| Parameters | Unit | Description |
|---|---|---|
| $q_i$ | degree | current position of joint i |
| $q^U_i$ | degree | upper limit of joint i |
| $q^L_i$ | degree | lower limit of joint i |
| $q^N_i$ | degree | neutral position of joint i |
| G | - | constant, $10^6$ |
| $QU_i$ | - | penalty term of upper limits |
| $QL_i$ | - | penalty term of lower limits |
| $\gamma_i$ | - | weighting value of joint i |

**Table 7: Parameters used in VSR discomfort index**

$$f(q)_{overall} = w_1 \frac{f_{fatigue}}{\max(f_{fatigue})} + w_2 \frac{f_{discomfort}}{\max(f_{discomfort})} \quad (10)$$

$$f(q)_{fatigue} = \sum_1^n \left(\frac{\Gamma_i}{\Gamma_{max}}\right)^2 \quad (11)$$

$$f(q)_{discomfort} = \frac{1}{G}\left[\gamma_i \left(\Delta q_i^{norm}\right)^2 + G\,QU_i + G\,QL_i\right] \quad (12)$$

$$\Delta q_i^{norm} = \frac{q_i - q_i^N}{q_i^U - q_i^L} \quad (13)$$

$$QU_i = \left(\frac{1}{2}\sin\left(\frac{5\left(q_i^U - q_i\right)}{q_i^U - q_i^L} + \frac{\pi}{2}\right) + 1\right)^{100} \quad (14)$$

$$QL_i = \left(\frac{1}{2}\sin\left(\frac{5\left(q_i - q_i^L\right)}{q_i^U - q_i^L} + \frac{\pi}{2}\right) + 1\right)^{100} \quad (15)$$

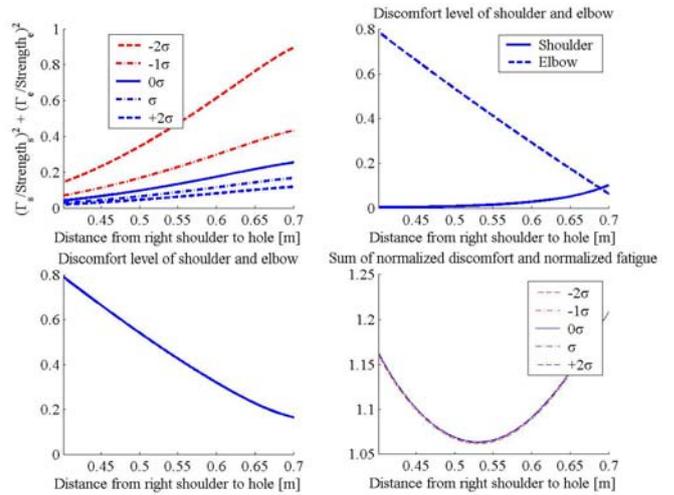

**Figure 12: Evaluation of the influence of the working distance**

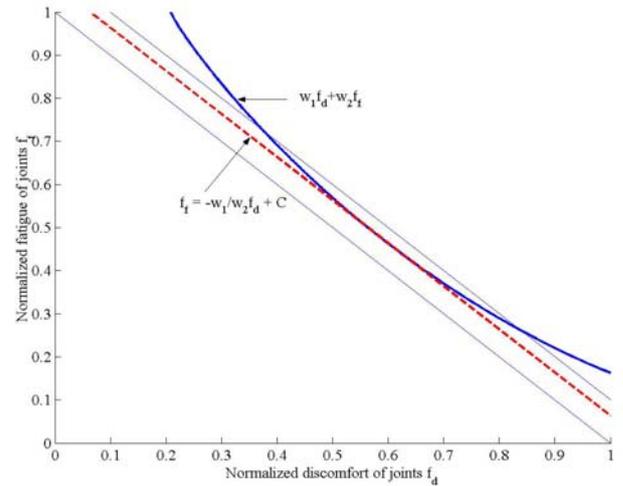

**Figure 13: Optimal posture analysis of the weighting values**

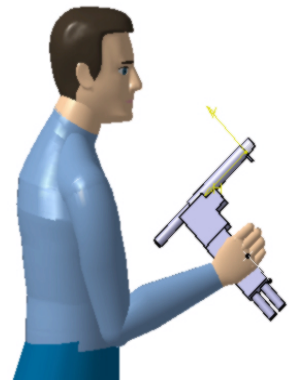

**Figure 14: Graphical visualization of the optimal posture**

In Figure 12, the left upper subfigure is the stress index of the 95% population. This stress index can represent certain fatigue level of a posture. The longer the distance, the larger moment arm of external loads, more stress it is for the right arm. The discomfort of right shoulder and elbow are shown in right upper subfigure. It is clear that the elbow becomes more comfort while the distance gets longer, since it approaches to its neutral position. Conversely, the shoulder gets more discomfort since it moves further from its neutral position. The sum of both discomforts is shown in left lower subfigure. After normalization of the stress and discomfort, both are added together to be a overall objective function for the optimization of the posture. Both factors are weighted by weight factor $w_i$. The variation of $w_i$ allows us to move the optimal solution along the Pareto surface in Figure 13. In our analysis, both weighting factors are set as 1. These weighting factors can be set according to the preference of less fatigue or less discomfort. In the right lower subfigure, an optimal posture can be found at the distance around 0.53m. The optimal position is graphically shown in Figure 14. The flexion angle of shoulder and elbow are 22° and 98° to maintain the drilling machine. By setting different weighting values, different optimal posture can be achieved.

5.2- Discussion

The main difference between the fatigue analysis in this paper and the conventional methods for posture analysis is: all the physical exposure factors are taken into consideration in this method as well, but in a continuous record method. In this way, much detailed analysis of the operation can be achieved.

With the new fatigue and recovery model, it is possible to evaluate the fatigue of a certain manual handling job. Although until now only a specific application case is analyzed, the feasibility of the overall concept is verified in this paper. The fatigue at each joint, the reduction of the joint strength and the recovery time necessary for preventing the worker from cumulative fatigue can be calculated out with respect to physical and temporal parameters of the job. With the analysis result, it is possible to determine suitable work-rest schedules to minimizing fatigue during a job, and to provide recommended postures to the user. With the analysis the distribution of the population, the fatigue model can also be used to select suitable workers for the jobs.

However, in a manual handling job, there are lots of objective factors influencing the performance of the worker, such as, temperature, vibration, and so on. In a virtual reality framework, it is impossible to reproduce all these factors. From another view, there are different aspects concerning the difficulty of the job, such as accessibility, visibility, comfort, and fatigue. In real working process, the worker can adjust the operation according to the environment, the requirement of the job and his own capacities. For this reason, the actual operation is the result of multiple-objective optimization. In the drilling case, multiple-objective optimization posture can be achieved by weighting fatigue and discomfort as the same. Although this cannot reflect the actual posture in the manual handling work, at least the result can provide us a recommended posture to decrease MSD risks.

## 6- Conclusions and Perspectives

In this paper, the application of a new muscle fatigue and recovery model in a virtual environment framework is presented. In the digital human simulation, the joint torque load can be calculated after geometrical and dynamic modelling of human. Thus, according to biomechanical limits of each joint, the fatigue level of the joint can be figured out using the fatigue model. Further more, the fatigue model and recovery model can be used to determine the work-rest schedule for manual handling jobs. Nevertheless, combining fatigue index and discomfort index of joint, virtual human's motion can be predicted or proposed in digital human simulation tools.

In the future, other manual handling jobs are going to be evaluated under this framework with consideration of fatigue. Full body geometrical and dynamic model of virtual human is going to be constructed to evaluate the joint fatigue for all the key joints of human. Experimental validation of the evaluation results is now under construction. It is possible to apply the new fatigue and recovery model in commercialized simulation software to simplify ergonomics evaluation procedures and enhance the work design efficiency, and make contribution to its final goal – reduce MSD risks in manual handling jobs.

## 7- Acknowledgements

This research was supported by the EADS and by the Région des Pays de la Loire (France) in the context of collaboration between the Ecole Centrale de Nantes (Nantes, France) and Tsinghua University (Beijing, P.R.China).

## 8- References

**[B1]** Burdorf A. Exposure assessment of risk factors for disorders of the back in occupational epidemiology. In Scandinavian Journal of Work, Environment and Health, 18: 1-9, 1992.

**[BP1]** Badler N. I., Phillips G. B. and Webber B. L. Simulating Humans: Computer Graphics Animation and Control, Oxford University Press, USA.1993.

**[C1]** Chaffin D. B. A computerized biomechanical model-development of and use in studying gross body actions. In Journal of Biomechanics, 2: 429-441, 1969.

**[C2]** Chaffin D. B.On Simulating Human Reach Motions for Ergonomics Analyses. In Human Factors and Ergonomics in Manufacturing, 12(3): 235-247, 2002

**[CA1]** Chaffin D. B., Andersson G. B. J. and Martin B. J. Occupational Biomechanics, Wiley-Interscience.1999.

**[CN1]** Carnahan B. J., Norman B. A. and Redfern M. S. Incorporating physical demand criteria into assembly line balancing. In IIE Transaction, 33: 875-887, 2001.

**[DH1]** Dan L., Hanson L. and O r. R. The Influence of virtual human model appearance on visual ergonomics posture evaluation. In Applied Ergonomics, 38: 713-722, 2007.

**[DW1]** Ding J., Wexler A. S. and Binder-Macleod S. A. A predictive model of fatigue in human skeletal muscles. In Journal of Applied Physiology, 89: 1322-1332, 2000.

**[DW2]** Ding J., Wexler A. S. and Binder-Macleod S. A. A Predictive Fatigue Model I: Predicting the Effect of Stimulation Frequency and Pattern on Fatigue. In IEEE Transactions on Neural Systems and Rehabilitation

Engineering, 10: 48-58, 2002.

**[DW3]** Ding J., Wexler A. S. and Binder-Macleod S. A. Mathematical models for fatigue minimization during functional electrical stimulation. In Electromyography Kinesiology, 13: 575-588, 2003.

**[EK1]** El ahrache K., Imbeau D. and Farbos B. Percentile values for determining maximum endurance times for static muscular work. In International Journal of Industrial Ergonomics, 36: 99-108, 2006.

**[FM1]** Forsman M., Hasson G.-A., Medbo L., Asterland P. and Engstorm T. A method for evaluation of manual work using synchronized video recordings and physiological measurements. In Applied Ergonomics, 33: 533-540., 2002.

**[FT1]** Freund J. and Takala E.-P. A Dynamic model of the forearm including fatigue. In Journal of Biomechanics, 34: 597-605, 2002.

**[GM1]** Giat Y., Mizrahi J. and Levy M. A Musculotendon Model of the Fatigue Profiles of Paralyzed Quadriceps Muscle Under FES. In IEEE Transaction on Biomechanical Engineering, 40: 664-674, 1993.

**[H1]** HSE. "Self-reported work-related illness in 2004/05." http://www.hse.gov.uk/statistics/swi/tables/0405/ulnind1.htm, 2005.

**[JJ1]** Jayaram U., Jayaram S., Shaikh I., Kim Y. and Palmer C. Introducing quantitative analysis methods into virtual environments for real-time and continuous ergonomic evaluations. In Computers In Industry, 57: 283-296, 2006.

**[KD1]** Khalil W. and Dombre E. Modelling, identification and control of robots, Hermes Science Publications.2002.

**[KK1]** Khalil W. and Kleinerfinger J. F. A new geometric notation for open and close-loop robots. In Proceedings of the IEEE Robotics and Automation: 1174-1180, 1986.

**[KS1]** Komura T., Shinagawa Y. and Kunii T. L. Calculation and Visualization of the Dynamic Ability of the Human Body. In Visualization and Computer Animation, 10: 57-78, 1999.

**[KS2]** Komura T., Shinagawa Y. and Kunii T. L. Creating and retargetting motion by the musculoskeletal human. In The Visual Computer, 16: 254-270, 2000.

**[LB1]** Li G. and Buckle P. Current techniques for assessing physical exposure to work-related musculoskeletal risks, with emphasis on posture-based methods. In Ergonomics, 42: 674-695, 1999.

**[LB2]** Liu J. Z., Brown R. W. and Yue G. H. A Dynamical Model of Muscle Activation, Fatigue, and Recovery. In Biophysical Journal, 82: 2344-2359, 2002.

**[MB1]** Ma L., Bennis F., Chablat D. and Zhang W. Framework for Dynamic Evaluation of Muscle Fatigue in Manual Handling Work. In IEEE International Conference on Industrial Technology, 2008.

**[MC1]** Ma L., Chablat D., Bennis F. and Zhang W. A New Dynamic Muscle Fatigue Model and its Validation, In International Journal of Industrial Ergonomics, 2008. doi: 10.1016/j.ergon.2008.04.004

**[ME1]** Mathiassen S. E. and Elizabeth A. Prediction of shoulder flexion endurance from personal factors. In International Journal of Industrial Ergonomics, 24: 315-329, 1999.

**[MR1]** Maier M. and Ross-Mota J. "Work-related Musculoskeletal Disorders." From http://www.cbs.state.or.us/ external/imd/rasums/resalert/msd.html, 2000.

**[R1]** Rodgers S. H. Recovery Time Needs for Repetitive Work. In Seminars in Occupational Medicine, 1986.

**[R2]** Rodgers S. H. Handbook of Human Factors and Ergonomics Methods, CRC Press: 12.11-12.10, 2004.

**[RB1]** Rodriguez I., Boulic R. and Meziat D. A Joint-level Model of Fatigue for the Postural Control of Virtual Humans. In Journal of Three Dimensional Images, 17: 70-75, 2003.

**[RB2]** Rodriguez I., Boulic R. and Meziat D. A model to assess fatigue at joint-level using the half-joint pair concept. In In Proceedings of the SPIE, Visualization and Data Analysis: 21-27, 2003.

**[RB3]** Rodriguez I., Boulic R. and Meziat D. Visualizing human fatigue at joint level with the half-joint pair concept. In In Proceedings of the SPIE, Visualization and Data Analysis: 37-45, 2003.

**[S1]** Safety and Health Assessment and Research for Prevention Work-related Musculoskeletal Disorders of the Neck, Back, and Upper Extremity in Washington State, 1995-2003, Washington State Department of Labor and Industries, 2005.

**[SL1]** Schaub K., Landau K., Menges R. and Grossmann K. A computer-aided tool for ergonomic workplace design and preventive health care. In Human Factors and Ergonomics in Manufacturing, 7: 269-304, 1997.

**[SH1]** Stanton N., Hedge A., Brookhuis K., Salas E. and Hendrick H. Handbook of Human Factors and Ergonomics Methods, CRC PRESS.2004.

**[V1]** Vignes R. M. Modeling Muscle fatigue in digital humans. Master thesis Graduate College of The University of Iowa, 2004.

**[V2]** Vollestad N. K. Measurement of human muscle fatigue. In Journal of Neuroscience Methods, 74: 219-227, 1997.

**[WD1]** Wexler A. S., Ding J. and Binder-Macleod S. A. A Mathematical model that predicts skeletal muscle force. In IEEE Transactions On Biomedical Engineering, 44: 337-348, 1997.

**[WF1]** Wood D. D., Fisher D. L. and Andres R. O. Minimizing Fatigue during Repetitive Jobs: Optimal Work-Rest Schedules. In Human Factors: The Journal of the Human Factors and Ergonomics Society, 39(1): 83--101, 1997.

**[YM1]** Yang J., Marler R. T., Kim H. and Arora J. S. Multi-objective optimization for upper body posture prediction. In 10th AIAA/ISSMO Multidisciplinary Analysis and Optimization Conference, 2004.